\documentclass{amsart}

\usepackage{amsmath}
\usepackage{amsthm}
\usepackage{amsfonts}
\usepackage{amssymb}
\usepackage{epsfig}
\usepackage{graphicx} 
\usepackage{palatino,url}

\theoremstyle{plain}

\theoremstyle{remark}

\theoremstyle{definition}

\include{psfig}

\DeclareMathOperator*{\argmax}{argmax}

\begin{document}
\title{Parallel AdaBoost Algorithm for Gabor Wavelet Selection in Face Recognition }
%
\author[Ba\u{g}c\i]{Ula\c{s} Ba\u{g}c\i}
\address{Collaborative Medical Image Analysis on Grid (CMIAG),The University of Nottingham, Nottingham, UK}
\email{ulasbagci@ieee.org}
\urladdr{www.ulasbagci.net}

\author[Bai]{Li Bai}

\begin{abstract}
In this paper, the problem of automatic Gabor wavelet selection for face recognition is tackled by introducing an automatic algorithm based on Parallel AdaBoosting method. Incorporating mutual information into the algorithm leads to the selection procedure not only based on classification accuracy but also on efficiency. Effective image features are selected by using properly chosen Gabor wavelets optimised with Parallel AdaBoost method and mutual information to get high recognition rates with low computational cost. Experiments are conducted using the well-known FERET face database. In proposed framework, memory and computation costs are reduced significantly and high classification accuracy is obtained.
\end{abstract}

\maketitle

\section{Introduction}
\label{sec:intro}
\noindent Automatic face recognition is a challenging problem with currently achievable levels of performance not adequate for universal practical application, and there remains a need for further work to improve performance and flexibility. Numerous algorithms have been developed  for face recognition since it has been proved that Gabor-type receptive field could extract the maximum information from local image regions~\cite{okajima} and Gabor filters function similarly to the visual neurons of the human visual system~\cite{daugman}. Therefore, mathematical transforms using Gabor wavelets (GW) play an increasingly important role in extracting robust features from face images for classification~\cite{shen,shen2,shen3,competition}.

Representing images by GW is difficult problem due to two reasons. First, since GW are not orthogonal,  they cannot be used as basis functions as the reconstruction coefficients will not be unique for each image. Second, although exploiting the locality property of GW allows convolution of a GW at each location of the image to extract detailed local image information, the application of GW for all possible orientations and scales at every location in the image results in an enormous computational overhead. High computational cost can be avoided by reducing the feature dimensionality which requires optimization of the criterions for selecting GW. 

Most existing research studies select Gabor wavelets empirically, rather than optimally. The challenge that researchers  are facing today is how best to exploit GW to maximize benefits in terms of object recognition performance. This paper aims to optimize the criterions for selecting GW by AdaBoost (AB) algorithm in a parallel manner and incorporating mutual information (MI) into the algorithm. After giving the theoretical background on GW in Section~\ref{sec:gw}, AB and Parallel AdaBoost (PAB) algorithms are explained in Section~\ref{sec:fext} as means of feature extraction and selection. Section~\ref{sec:MI} explains MI based GW selection procedures with AB and PAB algrotihms respectively. Experimental results in Section~\ref{sec:exp} are followed by  conclusion in Section~\ref{sec:conc}. 
\section{Gabor Wavelets (GW)}
\label{sec:gw}
In the spacial domain, the 2D Gabor filter is a Gaussian kernel modulated by a sinusoidal plane wave~\cite{shen,shen2}
\begin{equation}
\label{g1}
 \varphi_{\Pi(f,\theta,\gamma,\eta)}(x,y)=\frac{f^2}{\pi\gamma\eta}e^{-(\alpha^{2}x'^{2}+\beta^{2}y'^{2})}e^{j2\pi fx'},
\end{equation}
where $x'=xcos\theta + ysin\theta$, $y'=-xsin\theta + ycos\theta$, $f$ is the central frequency of the sinusoidal plane wave, $\theta$ is the anti-clockwise rotation of the Gaussian and the plane wave, $\alpha$ is the sharpness of the Gaussian along the major axis parallel to the wave, and $\beta$ is the sharpness of the Gaussian minor axis perpendicular to the wave.  $\gamma=\frac{f}{\alpha}$ and $\eta=\frac{f}{\beta}$ are defined to keep the ratio between frequency and sharpness constant. The Gabor filters, like many other wavelets, can be generated from one mother wavelet by dilation and rotation. Each filter is in the shape of plane waves with frequency $f$, restricted by a Gaussian envelope function with relative width $\alpha$ and $\beta$. To extract useful features from an image, normally a set of Gabor filters with different frequencies and orientations are required \cite{shen2,competition},
\begin{eqnarray}
 \label{g2}
 \varphi_{u,v}=\varphi_{\Pi(f_u,\theta_v,\gamma,\eta)}, \nonumber\\
 f_u=f_{max}/\sqrt{2^u},  \qquad \theta_v = \frac{v}{V}\pi, \nonumber\\
 u=0,\ldots ,U-1,\qquad  v=0,\ldots ,V-1.
\end{eqnarray}
As shown in Eq (\ref{g1}) and (\ref{g2}), the following parameters need to be determined to design Gabor filters for feature extraction: the highest peak frequency $f_{max}$, the ratio between centre frequency and the sharpness of Gaussian major axis: $\gamma$ and minor axis: $\eta$, the number of scales $U$ and orientations $V$. See our previous studies~\cite{shen, shen2,competition} for further theoretical details on how to select these parameters .

\section{Feature Extraction and Selection Using GW}
\label{sec:fext}
The aim is to use GW to extract unique features uniformly across all images so that these features can be compared for face recognition. A common approach is to convolve each image with the same set of GW. The number of Gabor wavelets used for this varies with different applications, but usually 40 filters ($U$=5 scales and $V$=8 orientations) are chosen empirically for face recognition applications~\cite{shen,shen2,wiskott,competition}. Specifically, given a bank of 40 GW $\left\lbrace \varphi_{u,v}(x,y), u=0,\ldots,4, v=0,\ldots,7 \right\rbrace $, image features at different locations, frequencies and orientations can be extracted by convolving the image $I(x,y)$, locally, with the GW  $O^{I}_{u,v}(x,y)=|I*\varphi_{u,v}|(x,y) $. The feature set thus consists of the results of the local convolution of the image $I(x,y)$ with all of the 40 GW
\begin{equation}
S=\left\lbrace O^{I}_{u,v}(x,y): u \in \left\lbrace 0,...,4\right\rbrace, v \in \left\lbrace 0,...,7\right\rbrace  \right\rbrace. 
\end{equation}
A Gabor feature vector can be obtained by concatenating the rows (or columns) of $O^{I}_{u,v}(x,y)$ for all $u,v$ to represent the image: $G(I)= \mathcal{O}=\left(O^I_{0,0},O^I_{0,1},\ldots,O^I_{4,7}\right) $ where $G(.)$ is the Gabor feature extraction operation. As an example, taking an image of size 64 x 64, the Gabor feature vector will be of 64 x 64 x 5 x 8=163.840 dimensions, which is incredibly large. Due to the large number of convolution operations, the computation cost is also necessarily high. 

Instead of performing a convolution operation at every image location, using all the 40 GW, it is more sensible to select only the relevant GW to perform convolution with the image at appropriate positions. Two questions arise from this consideration: first, which wavelets should be used and, second, at which image locations. To fully appreciate the solution of these questions, we have developed an approach using AB algorithm to select GW based on not only location parameter of GW, but also orientation and frequency parameters of GW in~\cite{shen3,competition}. In this study, we improve this approach further by introducing the MI concept to the feature selection procedure and parallelizing the AB algorithm. 

\subsection{Parallel AdaBoost (PAB) Algorithm}
\label{subsec:padaboost}

Briefly, AB algorithm iteratively builds a trainable model $M$ using linear superposition of different realizations. The base model $M$ is re-trainable by using different weight combinations, $\textbf{w}=w_1,w_2,\ldots w_N$, where $N$ is number of samples~\cite{boost, merler}. After each training step, the weights are updated according to classification performance of the previous step over the training data. The weights of misclassified points, $y_i=\left\lbrace -1\right\rbrace$, are increased and weights of correctly classified points, $y_i=\left\lbrace +1\right\rbrace$, are decreased accordingly~\cite{boost}. Therefore, at each step there is an associated model $M_k$. The final hypothesis/model is the linear superposition of all these model instances.

The AB algorithm is computationally expensive. In particular, for any ''hard'' point, the distribution of the associated weights appears to converge, as the number of the steps of the AB algorithm grows to infinity, to a definite, stable distrubition~\cite{merler}. PAB aims to decrease the computational cost by approximating these asymptotic distributions. It is shown that weight parameters can be modelled well  by Gamma distributions of suitable parameters~\cite{collins}.  Using early estimates of weights, one can construct a distribution system from which AB weights can be selected instead of waiting for the sequential outputs of each steps. Once weight distributions $\gamma_i^{*}$ are modelled under the Gamma distribution by 
\begin{equation}
 \gamma = \frac{x^{\alpha -1}e^{-x/\theta}}{\Gamma(\alpha)\theta^{\alpha}},
\end{equation}
then weights are updated independently and randomly from this distribution where values for $\alpha$ and $\theta$ are obtained from the mean, $\mu$, and the variance, $\sigma^2$, of the weights based on first $S$-step evolutions.  The relationship of these variables is the following:
 \begin{equation}
   \mu=\alpha\theta  \qquad \text{and} \qquad \sigma^2=\alpha\theta^2
\end{equation}

\subsubsection{P-Boost Algorithm}
Given the data set $E \equiv \left\lbrace \left( x_i,y_i\right) \right\rbrace_{i=1}^N$ ;
\begin{enumerate}
\item Initialize weights  $w_i(1) = 1/N, i=1,...,N$
\item Run AdaBoost for $S$ steps and keep weights for each step, $w_i(n), n=1,...,S$
\item For $i=1,...N$, estimate the distribution $\gamma_i^{*}$ from weights stored previously, $w_i(n)$.
\item PARALLEL COMPUTATION STARTS HERE\\
      For each value of $n \in\left\lbrace  S+1,...,T\right\rbrace $ : do the steps below in parallel  
\begin{enumerate}
\item For i running on the data set, generate random and independent weights $w_i^{*}(n)$ by sampling the corresponding $\gamma_i^{*}$;
\item  Train base model $M$ using weights $w_i^{*}(n)$, resultant model instance $M_n$;
\item  Compute model error $\epsilon_n$;
\item  Compute model weights $c_n$: $c_n=\frac{1}{2}$ln$(\frac{1-\epsilon_n}{\epsilon_n})$
\end{enumerate} 
\item Compute the output Hypothesis \\
$H(x)=\sum_{n=1}^{T} c_{n}M_{n}(x)$
\end{enumerate}
As easily seen that after the step 3, new values to the weights could then be assigned not by following the standard AB algorithm, but by randomly and independently sampling the respective Gamma distribution model. This leads dramatic reduction in computational cost without losing accuracy in classification performance due to correctly keeping dynamics of  stochastic process.

\subsection{Selecting Gabor Wavelets Using PAB}
We simplify the task of selecting GW for feature extraction from a multi-class face recognition problem to a two-class problem: selecting GW that are effective for intra- and extra-person space discrimination. Such selected GW should be robust for face recognition, as intra- and extra-person space discrimination is one of the major difficulties in face recognition. 

The transition from a multi-class to a two-class problem is based on a method proposed in~\cite{svm}, reformulating the face recognition problem as a two class problem. Two spaces, intra- and extra-person spaces are defined, with intra-person space measuring respectively dissimilarities between faces of the same person and extra-person space dissimilarities between different people. We define intra- and extra-person spaces as
\begin{eqnarray}
Intra=\left\lbrace |G(I_p) - G(I_q)|, I_p \sim I_q \right\rbrace \nonumber \\
Extra=\left\lbrace |G(I_p) - G(I_q)|, I_p \nsim I_q \right\rbrace,
\end{eqnarray}
where $I_p$ and $I_q$ are the facial images of persons $p$  and $q$ respectively. Now it is seen that intra- and extra person space discrimination is a two-class problem and to use PAB algorithm for selecting GW, the training set will be $\mathcal{IE}=Intra \cup Extra$. Samples in the intra-person space are regarded as positive examples whilst those from extra-person space are regarded as negative examples. Each weak classifier can be defined on one Gabor wavelet, such that the weak classifier determines the class of a vector based on a feature extracted from the vector using just this one Gabor wavelet. Selected weak classifiers (and therefore the corresponding GW) are therefore effective in discriminating intra- and extra-person classes, and should be used to extract features for face recognition. Recall that each component of a vector in $\mathcal{IE}$ is associated with a Gabor wavelet, i.e., it is obtained by convolving an image with a Gabor wavelet $f_j(I)=||G(I_p)-G(I_q)||_j$, therefore, a weak classifier can be defined as a simple threshold function on a component of the vector as
\begin{equation}
h_j = 
\begin{cases}
-1, & \text{if $ f_j(I) < \lambda_{j} $}\\
1, & \text{if $ f_j(I) \geq \lambda_{j} $,}
\end{cases}
\end{equation}
where $\lambda_{j}$ can be determined by the intra-person sample mean and extra-person sample mean
\begin{equation}
\label{eq:meanintraextra}
\lambda_{j} = \frac{1}{2}\left(\frac{1}{m}\sum_{p=1}^m\left((x_p)_j|y_p=1 \right) + \frac{1}{l}\sum_{q=1}^l\left((x_q)_j|y_q=-1 \right) \right), 
\end{equation}
where $m$ and $l$ are the numbers of intra- and extra-person samples, respectively.

In each of the PAB and/or AB iterations, the space of all possible weak classifiers is searched exhaustively to find the best weak classifier that will produce the lowest classification error. The error is then used to update the weights such that the wrongly classified samples get more focus. The resulting strong classifier is a weighted linear combination of all the selected weak classifiers. The PAB and/or AB algorithm select hundreds of features and weak classifiers to form the final strong classifier.

\section{Mutual Information Usage in Boosting Algorithms}
\label{sec:MI}
The PAB and AB algorithm select only features that perform ''individually'' best, and the redundancy among selected features is not considered. To eliminate redundancy, MI can be used. Before a new weak classifier is selected, the MI between the new classifier and those already selected is examined to make sure that the information carried by the new classifier has not been captured before. At stage $T$ where $T-1$ weak classifiers $\left\lbrace h_{v(1)},h_{v(2)},\ldots,h_{v(T-1)},\right\rbrace $ are selected, the function to measure the MI between a candidate classifier $h_j$ and the selected classifiers can be defined as follows
\begin{equation}
M(h_j)=\argmax_{t} I(h_j, h_{v(t)}) \quad t=1,2,\ldots,T-1.
\end{equation}

Each weak classifier is now considered as a random variable. The estimation of MI between two such variables, e.g. $r_1$ and   $r_2$, requires information about the marginal distribution $p(r_1)$, $ p(r_2)$ and the joint probability distribution $p(r_1,r_2)$, where $p(.)$ represents probability. Though a Gaussian distribution could be assumed, many of the features might not be Gaussian. To reduce the complexity and computation cost of the feature selection process, we therefore focus on binary random variables only, i.e. $r_1 \in \left\lbrace -1,+1\right\rbrace$, $r_2 \in \left\lbrace -1,+1\right\rbrace$. For binary random variables, the probabilities could be estimated by simply counting the number of possible cases and dividing that number by the total number of training samples. The value of $M(h_j)$ can be directly used to determine whether the new classifier is redundant or not. The value is compared with a pre-defined threshold $\delta^{MI}$, if it is bigger than the $\delta^{MI}$, we can deduce that the information carried by the classifier has already been captured. Besides MI, the classification error of the weak classifier is also taken into consideration, i.e., only those classifiers with small classification errors are selected. The features thus selected are uncorrelated with each other and are therefore non-redundant. 

\begin{figure}[h]
\begin{minipage}[b]{1.0\linewidth}
 \centerline{\epsfig{figure=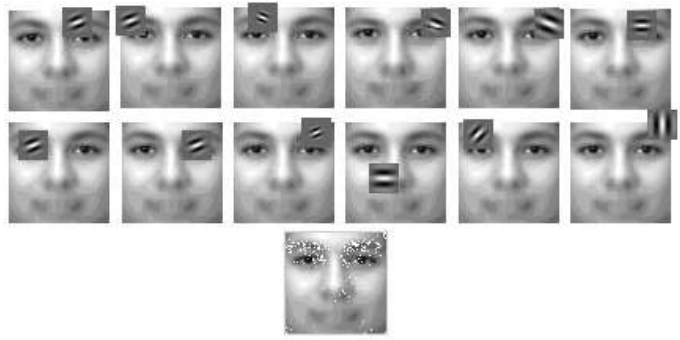,width=9cm}}
\caption{First 6, last 6, and position of 200 selected wavelets\label{img:face}}
\end{minipage}
\end{figure}

Fig.~\ref{img:face} shows the first and last six selected GW using MI enhanced PAB algorithm. It is interesting to see that most of the selected Gabor features are located around the prominent facial features such as eyebrows, eyes, nose and chin, which indicates that these regions are more robust against the variance of expression and illumination encountered within the database subset. This result is consistent with the fact that the eye and eyebrow regions remain relatively stable when a person's facial expression changes. Recall that the selection criterion is the ability of the GW in discriminating intra- and extra-person classes. 

\section{Experiments and Results}
\label{sec:exp}
We use a subset of 600 images from the FERET database to test the Gabor feature selection algorithm using AB, PAB and MI. Two images of each subject are randomly chosen for training, and the remaining one is used for testing. The selected 400 face images (2 images for each subject) are first used in boosting algorithms (AB and PAB) training to select GW for intra- and extra-person space discrimination. As a result, 200 intra-person difference samples and 1,600 extra-person difference samples are randomly generated for training~\cite{shen3}.

Although the required training time is longer than using the original AB due to the use of MI, the computational cost is reduced by using PAB algorithm so that required training time using MI with PAB is always lower than that of AB with MI. If the computational cost of AB algorithm is $\mathbf{O}(T)$, on the other hand, the cost of PAB algorithm is $\mathbf{O}(S)+(T-S).\mathbf{O}(1)$, where number of serial iterations $S$ in PAB is chosen smaller than total number of iterations $T$ in AB, $S<T$.

The normalized correlation distance measure and the nearest neighbor classifier are used. Table~\ref{table:results} shows the recognition performance on the 200 test images, where the highest accuracies achieved for the three algorithms are 93\%, 95\% and 96\% for AB, AB+MI and PAB+MI respectively. Since the MI values for all of the first 60 features are quite small, the effect of mutual information on the selection process is not obvious initially. However, once the number of features increases, AB and PAB start to pick up highly redundant features while the use of mutual information reduces the redundancy and improves recognition rate. In PAB, first 50 iterations are processed as AB, then the algorithm is parallelized in which weights in each iteration are selected randomly and independently from the model built using first 50 weights dynamics. To compare AB with PAB, not only computational cost is reduced dramatically, but also PAB algorithm appears to converge quickly to the reference model. 

\begin{table}[h]
\caption{Face Recognition Rates (\%) for various dimensions of feature set. AB: AdaBoost, AB+MI: AdaBoost with Mutual Information and PAB+MI: Parallel-AdaBoost with Mutual Information. \label{table:results}}

\begin{center}
\begin{tabular}{|c|c|c|c|}
\hline
Feature Dimension & AB  & AB+MI & PAB+MI, $S$=50\\ \hline 
20                         & 77.5 &77.5&77.5 \\  \hline
40                         & 82.0   &82.0 &82.0 \\  \hline
60                         & 86.0   &86.0 &86.0 \\  \hline
80                         & 87.5 &91.5 &91.5 \\  \hline
100                       & 91.0   &92.5 &93.5 \\  \hline
120                       & 92.0  &93.5 & 96.0\\  \hline
140                       & 93.0  &94.5 &96.0\\  \hline
160                       & 93.0  &93.5 &95.5 \\  \hline
180                       & 92.5 &95.0 &94.5 \\  \hline
200                       &92.5 &93.5 &93.0 \\  \hline
\end{tabular}
\end{center}
\end{table}

Table~\ref{table:results2} shows the recognition rates of PAB+MI algorithm for different values of $S$. Note that for all values of $S \in\left\lbrace < T=200\right\rbrace $, recognition rates are slighthly higher than the AB+MI case together with less computational cost respectively. The results indicate that PAB+MI method for various values of $S$ achieves the best result and converge quickly with respect to AB+MI case. Note that a few sequential steps are sufficient for PAB+MI to attain performances comparable with the reference AB+MI showing that weights dynamics are kept well with Gamma distribution.

\begin{table}[h]
\caption{PAB+MI Recognition Rates for different values of serial iteration number $n$ \label{table:results2}}

\begin{center}
\begin{tabular}{|c|c|c|c|c|}
\hline
Feature Dimension & $S$=50 & $S$=70 & $S$=100 & $S$=150\\ \hline 
20 &77.5 &77.5 &77.5 &77.5\\  \hline
40 &82.0 &82.0 &82.0  &82.0 \\  \hline
60 &86.0 &86.0 &86.0  &86.0 \\  \hline
80 &91.5 &90.0 &91.5 &91.5\\  \hline
100 &93.5 &91.5 &92.5 &92.5\\  \hline
120 &96.0 &92.5 &93.5 &93.5\\  \hline
140 &96.0 &96.0 &95.5 &94.5\\  \hline
160 &95.5 &94.5 &96.0 &93.5\\  \hline
180 &94.5 &96.0 &95.0 &94.0\\  \hline
200 &93.0 &95.5 &94.5 &91.0\\  \hline
\end{tabular}
\end{center}
\end{table}

\section{Conclusion}
\label{sec:conc}
The locality property of GW has both advantages and disadvantages.  A positive aspect is that it allows the extraction of local features, while a more negative aspect is its computational complexity due to uncertainty in the parameter selection process.
In this paper, we have discussed the effect of GW parameters on face recogniton performance and selection of GW for face recognition. We have introduced, step by step, the development process of GW selection method optimised for face recognition. 
These developments have demonstrated very encouraging results when investigated in a practical scenario as applied to the FERET face database. Work such as that reported here is important in demonstrating how PAB and MI techniques can be used to improve the effectiveness, efficiency and reliability of face recognition in biometrics-related applications.

\vfill
\pagebreak

\section{Acknowledgements}
This research is funded by the European Commission Fp6 Marie Curie Action Programme (MEST-CT-2005-021170) under the CMIAG (Collaborative Medical Image Analysis on Grid) project.


\end{document}